\documentclass{IOS-Book-Article}
\pdfoutput=1
\usepackage{mathptmx}

\usepackage[utf8]{inputenc} 
\RequirePackage{hyperref}
\RequirePackage{hypernat}
\usepackage{enumerate}
\usepackage{enumitem}
\usepackage{graphicx}
\usepackage{url}
\usepackage{amssymb}
\usepackage{enumerate}

\usepackage{algpseudocode, algorithm}

\usepackage{caption}
\usepackage{subcaption}
\usepackage{multirow}

\usepackage{4-charigruensenevomcguinness}

\def\hb{\hbox to 10.7 cm{}}

\begin{document}

\pagestyle{headings}
\def\thepage{}
\begin{frontmatter}  
\title{Directions for Explainable Knowledge-Enabled Systems}

\markboth{}{March 2020\hb}

\author[A]{\fnms{Shruthi} Chari}%
\author[B]{\fnms{Daniel M.} Gruen}
\author[A]{\fnms{Oshani} Seneviratne}
and
\author[A]{\fnms{Deborah} L. McGuinness}

\runningauthor{S. Chari, D. Gruen, O. Seneviratne, D. McGuinness}
\address[A]{Rensselaer Polytechnic Institute, Troy, NY, USA}
\address[B]{IBM Research, Cambridge, MA, USA}

\begin{abstract}
Interest in the field of Explainable Artificial Intelligence has been growing for decades, and has accelerated recently.
As Artificial Intelligence models have become more complex, and often more opaque, with the incorporation of complex machine learning techniques, explainability has become more critical.
Recently, researchers have been investigating and tackling explainability with a user-centric focus, 
looking for explanations to consider trustworthiness, comprehensibility, explicit provenance, and context-awareness.
In this chapter, we leverage our survey of explanation literature in Artificial Intelligence and closely related fields and use these past efforts to generate a set of explanation types that we feel reflect the expanded needs of explanation for today's artificial intelligence applications. We define each type and provide an example question that would motivate the need for this style of explanation. We believe this set of explanation types will help future system designers in their generation and prioritization of requirements and further help generate explanations that are better aligned to users' and situational needs.

\end{abstract}

\end{frontmatter}

\section{Introduction} \label{chapter41-introduction}
The field of Artificial Intelligence (AI) has evolved from solely symbolic- and logic-based expert systems to hybrid systems that employ both statistical and logical reasoning techniques. This shift and a greater incorporation of AI capabilities in systems across industries and consumer applications, including those that have significant, even life-or-death, implications have led to an increased demand for explainability. Advances in explainable AI have been tightly coupled with the development of AI approaches, such as the categories we covered in our earlier chapter, ``Foundations of Explainable Knowledge-enabled Systems," spanning expert systems, semantic web approaches, cognitive assistants, and machine learning methods. We note that these approaches tackle specific aspects of explainability. For example, explanations generated by expert systems and semantic applications primarily served the purposes for providing reasoning traces, provenance, and justifications. Those provided by cognitive assistants were capable of adapting their form to suit the users' needs, and, in the ML and expert systems domains, explanations provided an intuition for the model's functioning.

However, with the increased complexity of AI models, researchers have realized that the mechanistic explanation of the system's working alone might be insufficient for the end-users' needs. In a recent essay, Paez \cite{paez2019pragmatic} reasons that explanations need to convey a ``pragmatic and naturalistic account of understanding in AI." 
This idea of greater comprehensibility and user-focus is supported by several recent survey papers \cite{biran2017explanation, gilpin2018explaining, lipton2018mythos} and position statements \cite{mittelstadt2019explaining, doshi2017accountability}. 
In our earlier chapter, ``Foundations of Explainable Knowledge-enabled Systems," we presented definitions that we synthesized from the literature for explanations and explainable knowledge-enabled systems.

\subsection{Definitions}
\subsubsection{Explanation}
``An account of the system, its workings, the  \textit{implicit and explicit} knowledge it uses to arrive at conclusions in general and the specific decision at hand, that is \textit{sensitive} to the end-user's \textit{understanding, context, and current needs}."

\subsubsection{Explainable Knowledge-enabled systems}
``\textit AI systems that include a representation of the domain knowledge in the field of application, have mechanisms to incorporate the \textit{users' context}, are \textit{interpretable}, and host \textit{explanation facilities} that generate \textit{user-comprehensible, context-aware, and provenance-enabled} explanations of the mechanistic functioning of the AI system and the knowledge used."

\subsection{Overview}
In this chapter, we identify directions for research that could be instrumental in contributing to improving user aspects of explainable AI, providing explanations conducive ``to the end user's \textit{understanding, context, and current needs}," as previously described.  Additionally, we survey different explanation types that possess components, and exhibit presentation styles, tailored and variably suited for different contexts and situations. In Section \ref{chapter41-hybridexplanations}, we present a detailed overview of the explanation types that we have identified from the literature while focusing on their strengths and suitability to different AI scenarios. Further, in Section \ref{chapter41-directions}, we provide descriptions of directions for research that we believe will help generate various aspects of explainability, such as those related to causality and trustworthiness. In the same section, we also review methods that will help us better understand the explainability space, such as semantic representations and neuro-symbolic techniques. Ultimately, through our reviews in this chapter, we would like to highlight the idea that explanations are diverse, but always contain knowledge (model-specific, background, scientific, everyday, etc.) that can be variably presented in different presentation styles, and with different granularities, to suit the users' contexts, situations, preferences, and needs. 

\section{Hybrid Explanations} \label{chapter41-hybridexplanations}
As we discussed, explanations have evolved through shifts in the computing era.
As we suggested earlier in Section \ref{chapter41-introduction} and from our review of foundational approaches in our earlier chapter, we find that the generation of explanations have primarily been driven by capabilities of AI systems, and not by the demands of end-users. We identify that this is an issue, as consumers of AI systems, users reserve the right to understand and utilize results presented by the system they are using. Researchers \cite{doshi2017accountability, mittelstadt2019explaining, biran2017explanation} have noted that the users might not benefit from a mechanistic explanation of the system, and providing interpretable results alone is not sufficient for users to act on the conclusions produced by AI systems. Mittelstadt et al. \cite{mittelstadt2019explaining} and Biran and Cotton \cite{biran2017explanation} suggest that we need to look beyond the explanation types being generated by current AI systems and borrow from adjacent explanation sciences, such as social sciences and psychology. 

In our quest to develop explainable health assistants, as part of the Health Empowerment by Analytics, Learning, and Semantics (HEALS) project,\footnote{HEALS press release: \url{https://science.rpi.edu/biology/news/ibm-and-rensselaer-team-research-chronic-diseases-cognitive-computing}} we conducted a literature review to catalog the different explanation types. In Table \ref{tab01:explanationtypes}, we present our definitions of the nine explanation types that we researched in the form of a taxonomy. While our cataloging will eventually help us build a semantic understanding of the explanation space, it also helps us understand that each explanation type tackles different aspects of explainability, and we need to design hybrid explanations to meet the diverse needs of users, contexts, and situations. 

\begin{figure}[ht!]\centering
\includegraphics[width=1.0\textwidth]{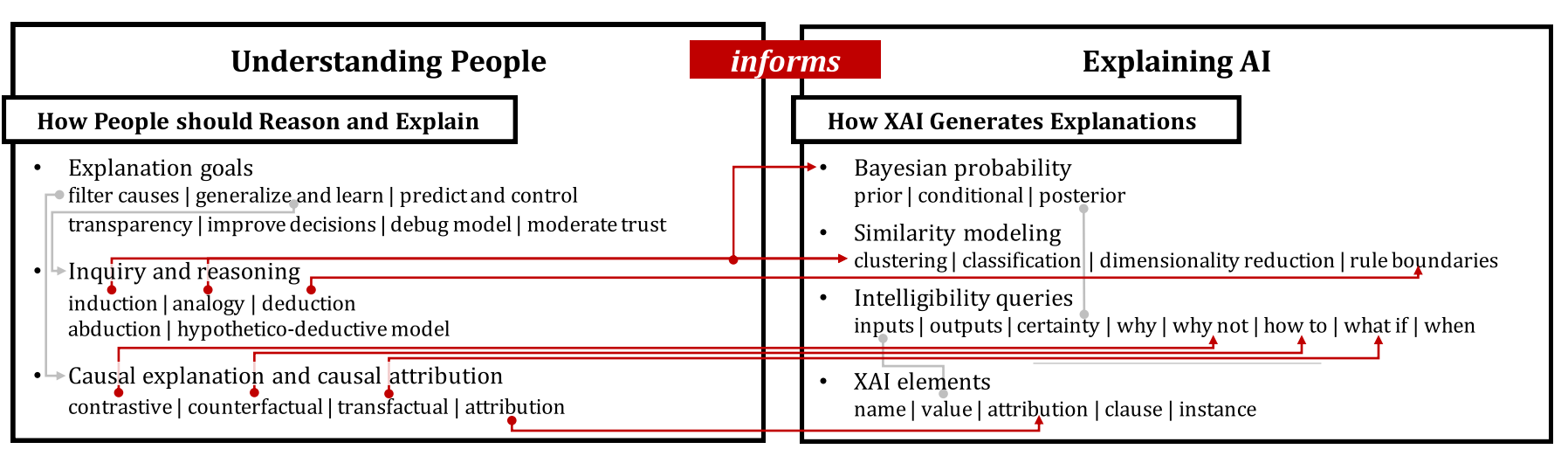}
\caption[limtaxonomyofintelligibility]{A partial conceptual framework mapping explanation types capable of being generated by explainable AI methods to the reasoning employed by users [Image taken from Lim et al. \cite{lim2019these} and Wang et al. \cite{ wang2019designing}]. For the full framework see Wang et al. \cite{wang2019designing}.}
\label{limtaxonomyofintelligbility}
\end{figure}

Similarly, a recent paper by Lim et al. \cite{lim2019these} investigates the link between user reasoning strategies and the reasoning strategies used by AI methods. This paper begins to connect how people reason and how AI methods can generate components that would satisfy the ``explanation goals" \footnote{The full list of explanation goals adapted from Nunes and Jannach \cite{nunes2017systematic} will be presented in Section \ref{chapter41-sec:semanticrep}} (e.g., improving decision-making, transparency, model debugging, etc.) that users desire. From an implementation standpoint, the authors build on an earlier taxonomy of intelligibility types proposed by Lim and Dey \cite{lim2009and}, and link intelligibility queries, including `Inputs,' `Outputs,' `Certainty,' `Why,' `Why Not,' `What If,' and `When,' to the associated explanation type being explored/generated by explainable AI models. Such a conceptual framework, as seen in Figure \ref{limtaxonomyofintelligbility}, is interesting, since they are beginning to think of the implementation of AI explainability from a user perspective. Additionally, the questions addressed by our identified explanation types, presented in Table \ref{tab01:explanationtypes}, are aligned with the intelligibility queries proposed in this paper, reiterating that each explanation type tackles different aspects of explainability. Efforts, such as this paper, that explore the diverse, explainable AI landscape, in conjunction with an understanding of the different explanation types, can help us generate explanations more suited to the users' needs.

In the rest of this section, we will present a review of the survey and position papers that we used as guides to present the explanation types. 
\begin{table}[!htbp]
    \caption{A catalog of different explanation types (ordered alphabetically), their definitions, and a motivating explanation question that a healthcare provider may ask.}
\label{tab01:explanationtypes}
    \centering
    \begin{tabularx}{\textwidth}{@{}lX@{}}
        \toprule
        \textbf{Explanation Type} & \textbf{Example Provider Question and  Literature-Derived Definition}\\
        \midrule
        Case-based & \textbf{``To what other situations has this recommendation been applied?"} Case-based explanations contain results that ``are based on actual prior cases that can be presented to the user to provide compelling support for the system’s conclusions" \cite{cunningham2003evaluation}. Borrowing from \cite{leake1988evaluating} and \cite{ahmed2015reasoning}, we opine that an AI system generating case-based explanations needs to remember and adapt explanations of similar prior cases \cite{leake1988evaluating}, or needs to reason ``from experiences (old cases) in an effort to solve problems, critique solutions and explain anomalous situations” \cite{ahmed2015reasoning}. Case-based explanations can involve analogical reasoning, relying on similarities between features of the case and of the current situation. \\ \hline
        Contextual & \textbf{``What broader information about the current situation prompted you to suggest this recommendation now?"} 
       Contextual explanations are those that refer to information about items other than the explicit inputs and output, such as information about the user, situation, and broader environment that affected the computation. Providing such information requires that a system be ``context-aware," and can include information about a ``user's tasks, significant user attributes, organizational environment, and technical and physical environments'' \cite{international1999human}.\\ \hline
        Contrastive & \textbf{``Why administer this new drug over the one I would typically prescribe?"} 
        As described by \cite{van2018contrastive} and \cite{miller2019explanation}, contrastive explanations define an output of interest and present contrasts between the fact (the event that did occur), the given output, and the foil (the event that did not occur), the output of interest. \\ \hline
        Counterfactual & \textbf{``What if the patient had a high risk for cardiovascular disease?  Would you still recommend the same treatment plan?”}
        Counterfactual explanations address the question of what results would have been obtained with a different set of inputs than those used. Paraphrasing \cite{woodward1997explanation}, counterfactual explanations are causal in nature and are generated by tracing patterns of a special kind of causal dependence. \\ \hline
        Everyday & \textbf{``Why are gloves recommended when dealing with high-risk patients?”} Everyday explanations are accounts of the real world that appeal to the user based on their general understanding and knowledge \cite{mcneill2008inquiry} of how the world works, and that help them understand why particular facts (events, properties, decisions, etc.) occurred \cite{miller2019explanation}.  There is evidence that users prefer everyday explanations that are causal in nature \cite{zemla2017evaluating}. \\ \hline
        Scientific & \textbf{``What is the biological basis for this recommendation?"} Scientific explanations reference the results of rigorous scientific methods, such as observations and measurements, to explain something we see in the natural world \cite{moore2000varieties}. Adapting from \cite{miller2019explanation}, we add that scientific explanations usually contain different components of interacting knowledge, including theories or mechanisms such as physiological ones, which are sets of principles that form building blocks for models; models which represent the relationships between entities and their attributes informed by taxonomies and other classification schemes; and data (e.g. measurements, observations).\\ \hline
        Simulation-based & \textbf{``What would happen if this recommendation is followed?"} Simulation-based explanations are those based on an imagined or implemented imitation of a system or process and the results that emerge from similar inputs.  As simulations can often be run numerous times (e.g. Monte Carlo simulations), and the mechanisms in the simulation can often be observed and traced directly, simulation-based explanations can have  elements of statistical and trace-based explanations. Heal suggests that these explanations \cite{heal1996simulation} contain facts that humans would use to determine an outcome in a specified case, and these explanations are intended to “replace and amplify real experiences with guided ones, often “immersive” in nature, that evoke or replicate substantial aspects of the real world in a fully interactive fashion" \cite{lateef2010simulation}. \\ \hline
        Statistical & \textbf{``What percentage of similar patients who received this treatment recovered?"} Statistical explanations present an account of the outcome based on data about the occurrence of events under specified (e.g., experimental) conditions. Statistical explanations refer to numerical evidence on the likelihood of factors or processes influencing the result. \cite{hempel1962deductive} add that a particularly high probability allows the outcome to be expected with practical certainty in any one case where the specified conditions occur.  \\ \hline
        Trace-based & \textbf{``What steps were taken by the system to generate this recommendation?"} Trace-based explanations describe the underlying sequence of steps used by the system to arrive at a specific result. They reveal ``the line of reasoning per case” \cite{lim2009and}, and ``addresses the question of why and how the application did something” \cite{lim2009and}. \\ 
        \bottomrule
    \end{tabularx}
\end{table}

\subsection{Findings from Review and Position papers}
A recent position paper \cite{mittelstadt2019explaining} presents links to explanations from the social sciences domain to the explanation needs in the explainable AI community. While this paper posits that their ideas are focused on how to better communicate the explanations of interpretability of black-box, deep learning models, the principles discussed apply to a broader class of AI models. The authors suggest that \textit{everyday} and \textit{scientific} explanations used in domains such as psychology and social sciences, are also applicable to explaining AI models because they are able to present abstract information at different granularities. Further, they draw a parallel between \textit{scientific} explanations and \textit{trace-based} explanations, in that both of these might not be understandable to all users. Hence, they present a case that, when explanations are delivered to humans who are selective and social in their processes, the explanations need to be \textit{contrastive} and \textit{communicative}. They conclude that the explainable AI community needs to provide explanations that are directly targeted and tailored to the needs of the users. 
We note that, in our taxonomy of explanation types (Table \ref{tab01:explanationtypes}), we don't account for implementation challenges, and we include certain explanations,  such as \textit{everyday} explanations, that require common-sense knowledge that may be difficult to gather and operationalize.  

Furthermore, another survey paper \cite{biran2017explanation} motivates the need to leverage 
explanation science literature from 
related fields, such as constraint programming, forensic sciences, context-aware systems, case-based reasoning, causal discovery, etc. The authors present a brief description of the usage of explanations in these domains, and use these related fields to motivate their plausible adoption in the explanation of the interpretability of ML models. In addition, the DARPA XAI report \cite{gunning2017explainable} also lists desirable explanation types, focusing on their form of delivery (explanation modalities), including visualizations, analytical statements, alternate choices, and case-based presentations.

\subsection{Summary}
Research in different explanation types is influenced by research from interdisciplinary fields, spanning social sciences and philosophy. While generating our catalog of definitions for different explanation types (Table \ref{tab01:explanationtypes}), we identified that certain explanation types, such as \textit{contrastive, counterfactual, case-based, and trace-based} explanations, are well-documented in the computer science literature. However, for some other explanation types, such as \textit{statistical, everyday, and scientific} explanations, we had to refer to literature from other domains, such as philosophy. We believe that explanations presented to users include components from different explanation types. For example, users require causal justifications to trust AI systems \cite{doshi2017towards}, statistical evidence for future exploration, scientific summaries to comprehend the system \cite{doran2017does}, and everyday explanations conducive to their understanding \cite{mittelstadt2019explaining}. Hence, we believe that there is a need to expand the explanation types our AI systems can currently support and build explanation facilities to generate hybrid and user-centric explanations. In Section \ref{chapter41-directions}, we present technologies that, in our opinion, will be instrumental in contributing to aspects of the various explanation components required to generate the explanation types shown in Table \ref{tab01:explanationtypes}.

\section{Directions} \label{chapter41-directions}
Today more than ever, there is a need to present \textit{personalized, trustworthy, and context-aware} explanations to users of AI systems \cite{mittelstadt2019explaining, biran2017explanation}. While defining explanations (Section \ref{chapter41-introduction}), we suggest that explanations should be generated with a user focus. This idea of keeping the end-user in mind while building explanation facilities is corroborated in \cite{mittelstadt2019explaining}, wherein they recommend that explanations should ``facilitate informed dialogue between users, developers, algorithmic systems, and other stakeholders." In this section, we seek to provide a review of approaches that we believe will be instrumental in increasing the user's trust in explanations and enabling more adaptive and user-centric explanations. The approaches that, in our opinion, will serve as the directions for research in explainable AI include Causal Methods, Neuro-Symbolic AI systems,
and representation techniques to model the explainability space and to enable trustworthy data sharing that includes nascent approaches, such as Distributed Ledgers Technology (DLT). We posit that causal methods that provide causal justifications for decisions will help in building the users' trust in the system \cite{lipton2018mythos}. On the other hand, Neuro-Symbolic approaches will improve the intelligibility issue\footnote{Our definition of intelligibility is very similar to the description proposed by Lipton \cite{lipton2018mythos} and Lou et al. \cite{lou2012intelligible}, in that intelligible models are interpretable wherein the contribution of model features to a decision can be deciphered.} aspect of ML models. Semantic representations of the explainability space will aid in systematically understanding and identifying aspects of explanations. Such organizations will then further help in building AI systems that will assist users via a ``Distributed Cognition" approach \cite{hollan2000distributed} where the system generates explanations aligned with the users' requirements. Furthermore,  as a technology that champions trustworthy interactions between mutually distrusting parties, DLTs are emerging as one of the many solutions to tackle trust issues in AI ``black-box" models, and address the lack of explainability by providing data and AI model provenance that cannot be repudiated.

\subsection{Causal Methods} \label{chapter41-subcausalmethods}
Causality has been explored as a critical component of explanations since at least the 1990s \cite{pearl2017theoretical, woodward2005making}. Causality and causal reasoning have been pursued as research domains often independently of ML and Semantic Web efforts. However, AI researchers are now realizing and starting to suggest that causality is vital for presenting explanations to end users \cite{doshi2017towards, ribeiro2016should, gilpin2018explaining}. Doshi and Kim \cite{doshi2017towards} cite causality as a
desired property for explainability. They state \cite{doshi2017towards} that causal reasoning can be used to explain when a ``predicted change in output due to a perturbation will occur in the real system.”
This idea that the system should respond to a causal dependence also suggests that the system should encode causal knowledge, which as per Pearl \cite{pearl2019seven}, 
is missing from association based AI methods.
\begin{figure}[ht!]\centering
\includegraphics[width=0.7\textwidth]{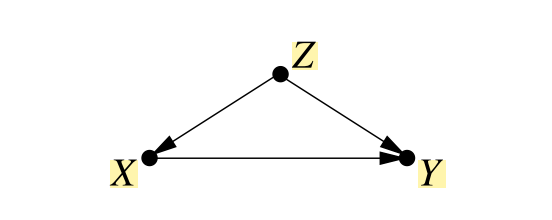}
\caption[A representation of Pearl's cause-effect model \cite{pearl2019seven, pearl2009causality} where $Q = P(Y|do(X))$, where $X$ has an effect on $Y$ and both depend on Z. Hence, he formulated the overall problem as a Bayesian equation in that $E_{z} = \sum_{z} P(Y|X, Z)P(Z)$. Pearl provides an intuitive example \cite{pearl2019seven} of gender ($Z$) being a confounder on the effect that taking a drug ($X$) will have on recovery ($Y$). [Image is taken from Pearl \cite{pearl2019seven} with permission from the author, Prof. Judea Pearl. ]{A representation of Pearl's cause-effect model \cite{pearl2019seven, pearl2009causality} where $Q = P(Y|do(X))$, where $X$ has an effect on $Y$ and both depend on Z. Hence, he formulated the overall problem as a Bayesian equation in that $E_{z} = \sum_{z} P(Y|X, Z)P(Z)$. Pearl provides an intuitive example \cite{pearl2019seven} of gender ($Z$) being a confounder on the effect that taking a drug ($X$) will have on recovery ($Y$). [Image is taken from \cite{pearl2019seven} with permission from the author, Prof. Judea Pearl. \protect \footnotemark ]}
\label{chapter41-docalculus}
\end{figure}

\footnotetext{Judea Pearl (judea@cs.ucla.edu) is a professor of computer science and statistics and director of the Cognitive Systems Laboratory at the University of California, Los Angeles, USA.}
A representation of causal models
has been presented by \cite{read1987constructing} 
and may be most simply presented as the well-known three-step Bayesian, cause-effect model proposed by Pearl \cite{pearl2009causality}. Read \cite{read1987constructing}, positioned his knowledge structure on Schank and Abelson's \cite{schank1975scripts} findings that humans cognitively inferred their next set of actions or exhibited a behavior based on a cause from an event. However, Pearl took his representation of causality a step further in introducing counterfactuals as a component. He noted that counterfactuals played a significant role in scientific and legal thinking, where the ``what if" and ``but if not for" type of questions are asked to identify the cause of a problem. 
Pearl has made significant contributions to the field of causality \cite{pearl2017theoretical, pearl2019seven, pearl2009causality}, and in our review of causal methods, we will summarize some of his most relevant 
contributions. Through our review of causal methods, 
we will show the need to 
more fully
integrate them
into future, hybrid AI approaches to address user-centric questions.

\begin{figure}[ht!]\centering
\includegraphics[width=1.0\textwidth]{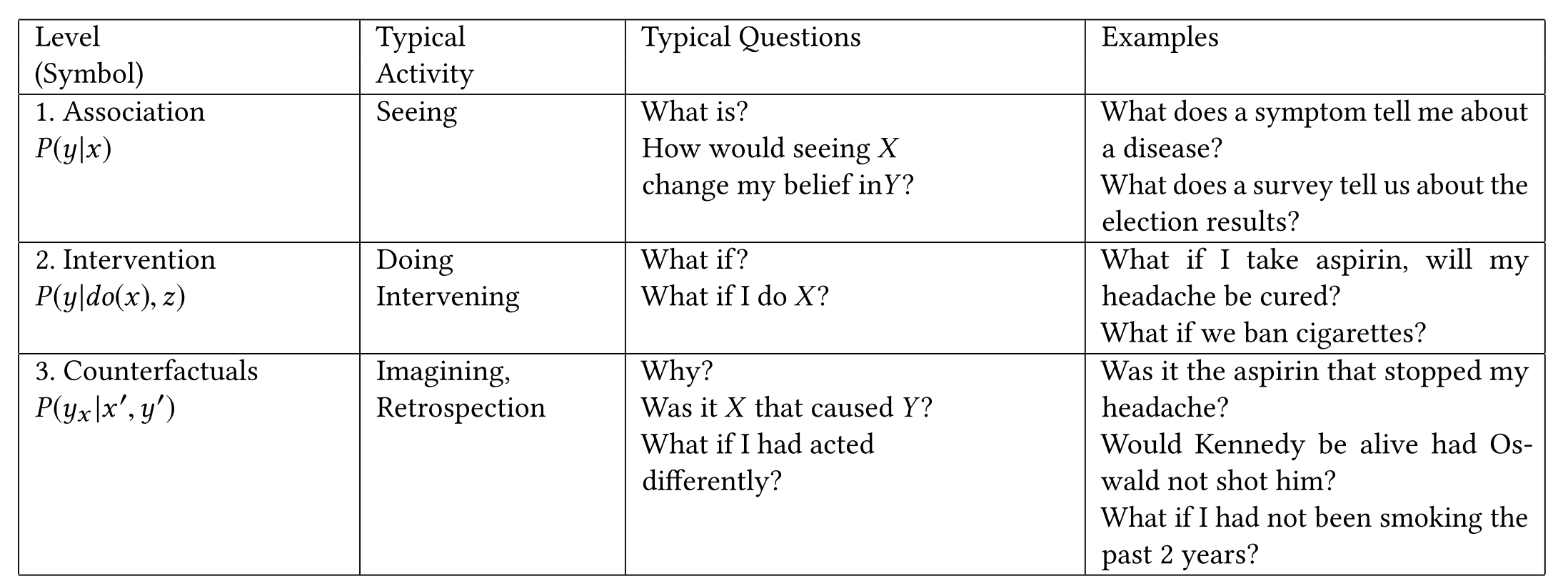}
\caption[pearldoquestions]{An organization of the typical questions and examples tackled by each level in the three-level causal hierarchy proposed by Pearl [Image is taken from Pearl  \cite{pearl2019seven} with permission from the author, Prof. Judea Pearl.] }

\label{pearldoquestions}
\end{figure}

In his widely-cited book \cite{pearl2009causality},
Pearl introduced a causal model for representing cause-effect relationships (Figure \ref{chapter41-docalculus}). This mathematical formulation of causality enabled researchers in fields, such as epidemiology and life sciences, to express causal structures \cite{pearl2018book}. In addition, one of his recent technical reports \cite{pearl2019seven} abstracts his cause-effect model and presents an overview of the three-step knowledge hierarchy (Figure \ref{pearldoquestions}) of causality that is comprised of Association, Intervention, and Counterfactual knowledge \cite{pearl2019seven}. Pearl notes that current ML techniques can address questions on \textit{Association} knowledge 
(i.e., Why am I being shown this answer? What else can I buy in addition to toothpaste?). In other words, \textit{Association} knowledge contains  correlations learned from associations. However, he adds that questions on \textit{Intervention} knowledge require the system to understand and encode knowledge about the world besides just the data it is inferring a decision on. Finally, he states that \textit{Counterfactual} questions that address the ``but why not" question would need the system to be aware or understand the cause-effect relationships. We believe that this clear separation and identification of knowledge, in a hierarchical fashion, would allow AI systems to identify the components that would be necessary to generate explanations for these broad knowledge categories. While causal structures are desirable, it is generally hard to discover these models due to their dependence on human cognition. However, there have been approaches that mimic human reasoning and identify causal relationships from text \cite{asghar2016automatic, girju2002text, barbey2007learning, kaplan1991knowledge}\footnote{We list cause-effect words from Pearl's report \cite{pearl2019seven}, but more can be found in the citations we have linked.} through the leveraging of the semantics of causal mentions. These techniques look for words such as the ones listed in Pearl's report \cite{pearl2019seven}, including ``cause," ``allow," ``preventing," ``attributed to," ``discriminating" and ``should I". Further, in the same report \cite{pearl2019seven}, Pearl presents seven tools in which causal methods are required:
\begin{enumerate}
    \item Encoding Causal Assumptions – Transparency and Testability 
    \item Do-calculus and the control of confounding
    \item The Algorithmization of Counterfactuals
    \item  Mediation Analysis and the Assessment of Direct and Indirect Effects
    \item  Adaptability, External Validity, and Sample Selection Bias
    \item Recovering from Missing Data
    \item Causal Discovery
\end{enumerate}

We believe that some of these tools, like Algorithmization of Counterfactuals, Causal Discovery, and Assessment of Direct and Indirect Effects,
will be particularly useful to include in explanations that provide the users' causal justifications for the conclusions being recommended to them by the AI system. 

In conclusion, we believe that causal representations will enable the ability of AI systems to address a broader class of explanations beyond the traditional ``Why, What, and How" \cite{dhaliwal1996use} questions. Additionally, with a concrete, cause-effect graphical model, such as the one proposed by Pearl \cite{pearl2018book, pearl2009causality}, 
the field has moved closer to a semantic representation of causality that may be used in a wide range of implemented systems.
Such a semantic representation of causal structures in KGs would lend to the development of causal, neuro-symbolic integrations. 

\subsection{Neuro-Symbolic AI Methods}
Neuro-Symbolic integration is a hybrid field that marries 
inductive and statistical learning capabilities of ML methods with the symbolic and conceptual representation  
capabilities of knowledge representation disciplines. 
Neuro-Symbolic Integration 
is not a new field 
\cite{bader2005dimensions, mooney1990experimental}, 
however, 
there has been a resurgence in interest due to its connection to explainable AI.
In this section, we discuss some opinions from the literature to demonstrate the capabilities of Neuro-Symbolic Integration.

In their position paper Hitzler et al. \cite{hitzler2019neural}, present distinctions between neural and Symbolic AI techniques to suggest that, with their contrasting strengths and applications, each of these two systems can assist each other to build a comprehensive solution. They point out that neural ML methods with their ``connectionist approach"\footnote{In the literature \cite{bader2005dimensions, mooney1990experimental}, connectionist approaches have been associated with neural networks that connect layers and nodes within layers.} are robust, noise-tolerant, and have the ability to identify patterns that even humans find hard to identify without prior training. On the other hand, the authors state that the semantic representation of knowledge allows symbolic AI methods to derive deeper relationships and provide high-confidence, provenance-aware results. However, they point out that the considerable reliance of symbolic AI methods on logical encoding makes them \textit{brittle} and less-tolerant towards data flaws and noise. Conversely, in-line with the position from other interpretable AI papers \cite{ribeiro2016should, lipton2018mythos}, they state that ML methods suffer from being unintelligible and have non-transparency issues. Hence, through presenting the strengths and weaknesses of neural and symbolic AI techniques, they affirm the need for Neuro-Symbolic Integration.  

\begin{figure}[ht!]\centering
\includegraphics[width=1.0\textwidth]{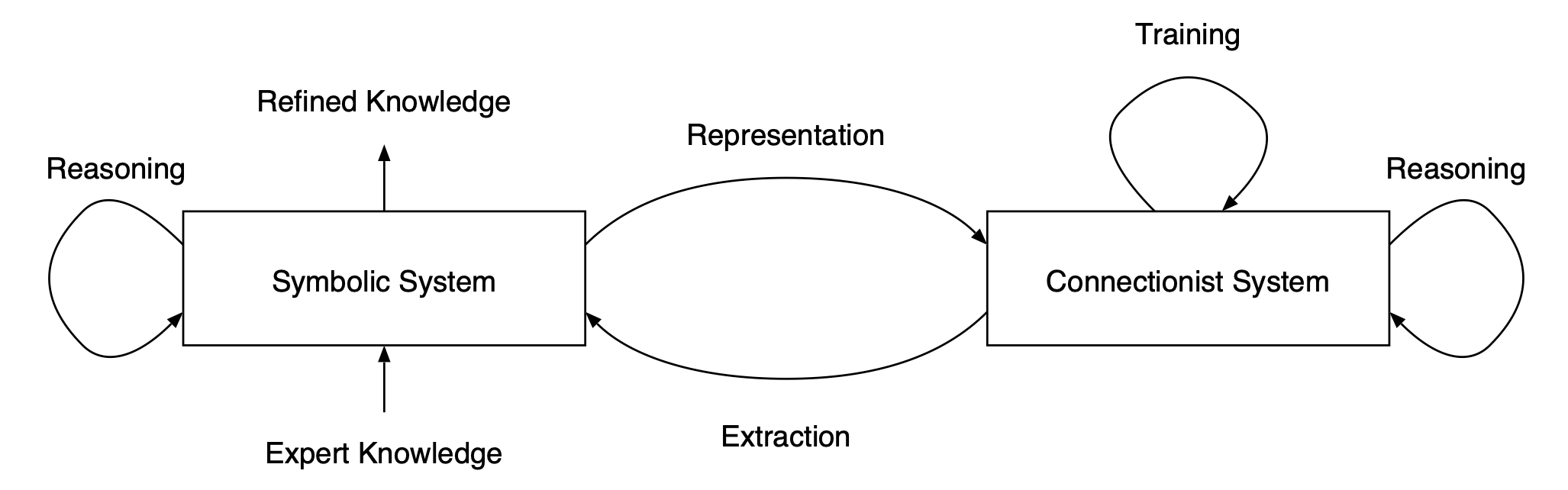}
\caption[neurosymbolic]{A schematic overview of a Neuro-Symbolic integration where connectionist approaches, such as ML methods and symoblic AI methods, help each other. [Image is taken from Bader and Hitzler \cite{bader2005dimensions}]}
\label{neurosymbolic}
\end{figure}

Further, Hitzler et al. \cite{hitzler2019neural}, identify tasks that will benefit from a Neuro-Symbolic Integration, including \textit{knowledge acquisition, fuzzy reasoning, and interpreting deep learning methods}. An illustration of a typical Neuro-Symbolic AI system where neural approaches aid symbolic systems in knowledge generation, and where symbolic systems provide the knowledge encoding to explain the functioning and results of neural methods, is seen in Figure \ref{neurosymbolic}. We believe each of the tasks identified in \cite{hitzler2019neural} will play a key role in the development of the hybrid knowledge-enabled systems (definition under Section \ref{chapter41-introduction}).
More specifically, a strong, scalable encoding of user and domain knowledge can help ML methods be ``transparent, understandable, verifiable, and trustworthy" \cite{hitzler2019neural}. 

A knowledge acquisition use case is demonstrated by Alshahrani et al. \cite{alshahrani2017neuro} in a biomedical setting, where they use neural methods to learn enriched KG embeddings of RDF and OWL representations of biomedical data and knowledge. The authors combine data from widely-used biomedical ontologies and knowledge bases, including the Gene Ontology \cite{ashburner2000gene}, the Human Phenotype Ontology \cite{robinson2008human}, the Side Effect Resource (SIDER) \cite{kuhn2016sider}, etc.\footnote{We only list a few of the knowledge bases used in Alsaharani et al. \cite{alshahrani2017neuro}, find the complete list in their paper} In this effort, they use a random walk algorithm to learn the local representations of KG nodes in such a manner so that deep learning models can utilize the semantic node rich content present in KGs. Through embeddings that allow for the combination of data and information, they find that traditional ML methods are capable of finding more drug-drug and drug-disease interactions. More recently, researchers from the University of Massachusetts, Amherst are working on Box embeddings of KGs \cite{vilnis2018probabilistic} to represent rich and fine-grained concepts present in KGs, such as transitive relations, definitions of negative properties, etc. Further, the authors allow for the representation of probabilistic scores in the Box embeddings to model uncertain knowledge. The KG embedding efforts are not only allowing neural approaches to leverage knowledge in their predictions, but are enriching semantic methods by allowing for the ability to draw inferences without relying solely on crafted inference rules.

In summary, while the role of Neuro-Symbolic Integration might not be directly observable in explanations produced by explainable AI, the capabilities enabled by this integration will allow for the inclusion of knowledge in ML methods. This combination of data and knowledge will help ML methods provide provenance-aware and grounded results. Additionally, this integration will help symbolic systems be probabilistic and fuzzy, allowing them to be dynamic to user requests. Hence, we believe that Neuro-Symbolic Integration is desirable for the development of knowledge-enabled systems.  

\subsection{Semantic Representation of the Explainability space} \label{chapter41-sec:semanticrep}
Since the emergence of Semantic Web \cite{hendler2002integrating} technologies and the renewed interest in explainability since the late 2000s, there have been few noteworthy efforts \cite{nunes2017systematic, tiddi2015ontology} to represent explanations and their dependencies in the AI world. These representation efforts have begun to result in the development of information artifacts, such as explanation taxonomies \cite{nunes2017systematic, lim2019these}, general-purpose knowledge graphs \cite{tiddi2014dedalo}, and ontologies \cite{tiddi2015ontology}. As we stated in Section \ref{chapter41-hybridexplanations}, building a semantic understanding of explanations will help us identify components that contribute to them, and will enable the development of hybrid AI models that are adept at generating them. In this section, we review a taxonomical structuring of explanations \cite{nunes2017systematic}, a knowledge graph framework \cite{tiddi2014dedalo} and ontology design pattern for explanations \cite{tiddi2015ontology}. 
The taxonomy and ontology design pattern for explanations generates a knowledge representation of explanations (with different granularities) upon the analysis of its dependencies, usage in different fields, and the goals that they support. Additionally, Dedalo, the knowledge graph framework \cite{tiddi2014dedalo}, provides a method to identify background knowledge for information clusters generated by AI models.

\begin{figure}[ht!]\centering
\includegraphics[width=1.0\textwidth]{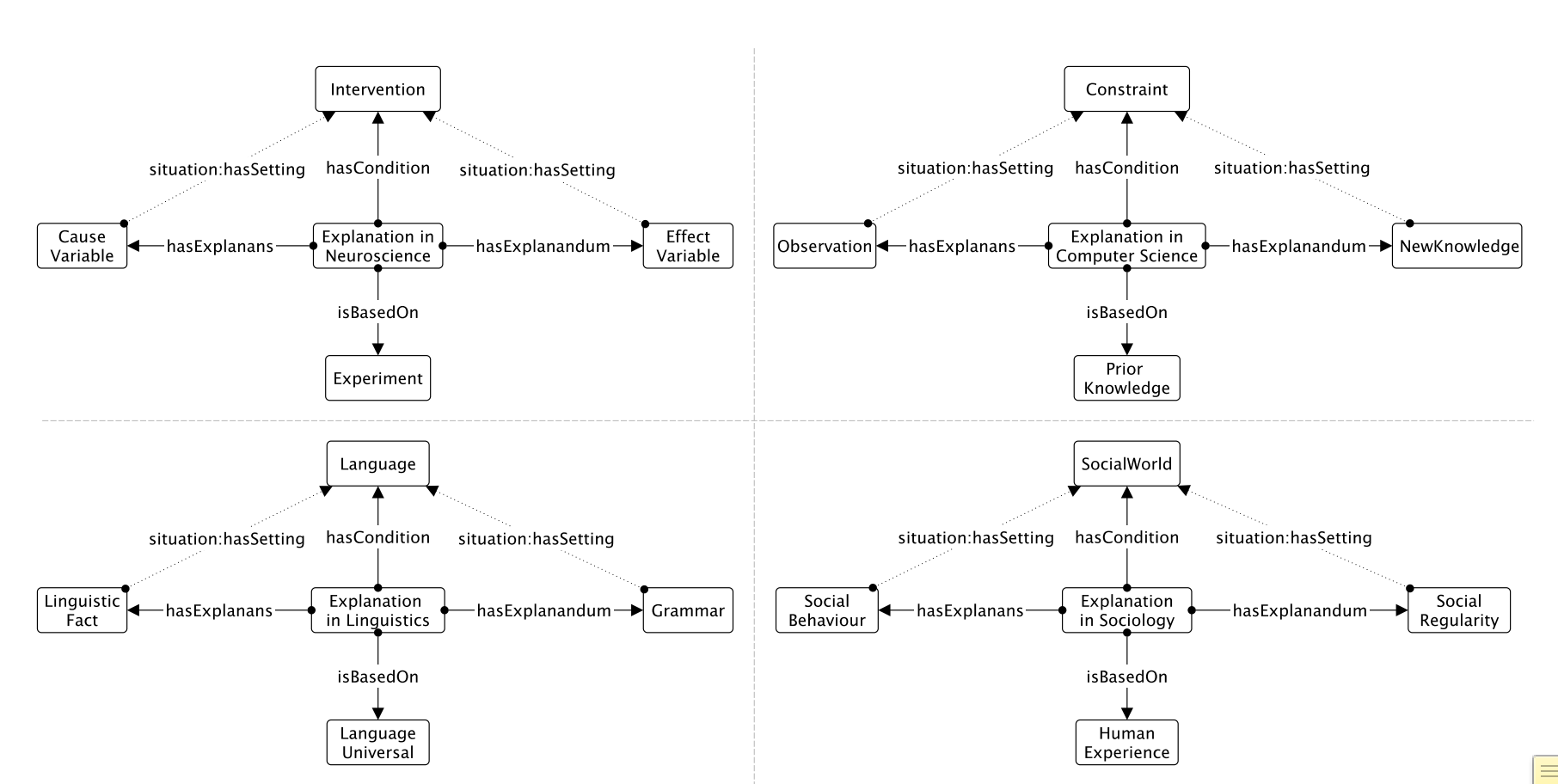}
\caption[ontologyexplanations]{Conceptualization of explanations in different fields based on the ontology design pattern proposed by Tiddi et al.\cite{tiddi2015ontology}. It is interesting to note that the authors have used the same property set to model explanations and their components in different fields. Such an encoding highlights that, while explanations serve diverse purposes and are instantiated by different components across fields, the structural composition remains the same in that explanations have `setting,' `condition,' and are based on some `theory' [Image is taken from Tiddi et al.  \cite{tiddi2015ontology}].}
\label{tiddiontologydifferenttypes}
\end{figure}

Tiddi et al. developed a design pattern for explanations upon surveying the role of explanations in various fields, spanning Linguistics, Computer Science, Neuroscience, and Sociology \cite{tiddi2015ontology}. While they found that the components of explanations vary in each field, they identified that certain components could minimally represent explanations. These components included the associated 
`event(s),' \footnote{Ontology classes labels are referred to in single quotes, and are in-line with the terminology used in the original paper \cite{tiddi2015ontology}} underlying `theory(ies),' `situation(s)' the explanations are applied to, and `condition(s)' the explanations utilize.
Additionally, in their ontology design pattern the authors incorporated standard nomenclature such as `explanandum' (``that which is explained") and `explanans' (``that which does the explaining") to associate explanations with the accounts of the premise that they are linked to and the strategies that are used to generate these accounts, respectively. With the explanation components and nomenclature in place, the authors solidified the representation of explanations as a quad, expressible as  $E $ = $\{ A, P, T, C \}$, ``where $A$ stands for the antecedent event/explanans, $P$ for the posterior event/explanandum, $C$ for the situational context they are happening in and $T$ for the theory governing those events." \cite{tiddi2015ontology} They used this backbone notation of $E $ = $\{ A, P, T, C \}$ to represent explanations in different fields (see Figure \ref{tiddiontologydifferenttypes}). An example would be in the Neuroscience domain where explanations are based on results from experiment ($T$), have a pre-event in a cause variable ($A$), and output a posterior result in the form of an effect variable ($P$).

In a related effort, Tiddi et al. developed the knowledge graph framework, Dedalo \cite{tiddi2014dedalo}, to extract background knowledge from Linked Data that can be used to populate explanations of clusters irrespective of the field of application. For this purpose, the authors used an Inductive Logic Programming (ILP) approach in conjunction with a heuristically driven method to identify background knowledge clauses to explain/interpret the results of AI models. A sample of an output fact from Dedalo could be ``Enrico Motta is associated with the Semantic Web." While Dedalo is a useful, general-purpose fact identifier, it might not be sufficient to entirely extract or identify knowledge necessary to generate all of the explanation types that we identified in Table \ref{tab01:explanationtypes}. Furthermore, neither Dedalo, nor the ontology design pattern for explanations, account for the different purposes that the explanations serve. However, this gap is somewhat addressed by the taxonomy of explanations developed by Nunes and Jannach \cite{nunes2017systematic}. 

More recently, Nunes and Jannach \cite{nunes2017systematic} conducted a systematic review of papers in explainable AI, making several analyses of the explanation space on the basis of their following research questions:
\begin{enumerate}
    \item What are the characteristics of explanations provided to users, in terms of content and presentation?
    \item How are explanations generated?
    \item How are explanations evaluated?
    \item What are the conclusions of evaluation or foundational studies of explanations?
\end{enumerate}
Their analyses of the various papers considered in their review resulted in interesting findings, including a catalog of explanation goals, and the categorization of different forms of knowledge that constitute the explanation components. The explanation goals spanned properties such as \textit{Transparency, Effectiveness, Trust, Persuasiveness, Satisfaction, Education, Scrutability, Efficiency}, and \textit{Debugging}. Furthermore, they grouped the knowledge into broad- and low-level categories, of which the broad-level categories were comprised of \textit{preferences and inputs, decision inference process, background information}, and \textit{alternative information}. Finally, the authors used their findings to construct a taxonomical organization of the explainability space, \footnote{For the full taxonomy of explanations diagram refer to Figure 11 in \cite{nunes2017systematic}} associating with explanations and components such as their \textit{level of detail}, \textit{objectives} (purpose, stakeholder goals, user-perceived factors), \textit{generality}, \textit{user interface components} (presentation and content), etc. While the ontology design pattern for explanations simplifies the model of explanations based on its content dependencies,  the taxonomy of explanations provides a comprehensive view of explanations with factors, such as the goals they address, the content they contain, the user purpose they serve, etc. 

In summary, the semantic representations of explanations, such as the ones we reviewed, can not only help provide a more precise understanding and organization of the explainable AI space but can also improve the flexibility of constructing AI models that serve the users' needs for explainability. Specifically, we believe that these semantic representations, that are being actively explored as a by-product to meet the growing needs of explainable AI \cite{mittelstadt2019explaining, doshi2017accountability, lipton2018mythos}, are a step towards generating hybrid explanations (such as the ones mentioned in Section \ref{chapter41-hybridexplanations}) with various strengths and capabilities. Further, while these representations might not contribute to techniques that generate explanations, they can aid in ensuring that the explanations generated are in-line with our definition of the explainable knowledge-enabled system, i.e., 
to generate explanations that are ``user-comprehensible, context-aware and provenance-enabled explanations of the mechanistic functioning of the AI system and the knowledge used."
\subsection{Distributed Ledger Technology for Knowledge-enabled AI}
\label{chapter41-sec:DLT}

In recent decades, one key priority in AI research has been pursuing optimal performance,
often at the expense of interpretability \cite{gunning2017explainable}. 
However, the crucial questions, driven by a social reluctance to accept AI-based decisions, may lead to entirely new dynamics and technologies, fostering explainability, and authenticity. 
Distributed Ledger Technology (DLT) is emerging as one of the many solutions to tackle trust issues in AI models, and addresses the lack of explainability by providing data and cryptographically verifiable AI model provenance. DLTs provide the following four key features that are desirable for explainable AI: 
\begin{enumerate}
    \item Transparency and visibility of the data and AI algorithms
    \item Immutability of the input data and parameters
    \item Traceability and nonrepudiation of the output
    \item Automatic execution of logic through smart contracts
\end{enumerate}

In cases where the data provider is concerned about data misuse, the DLT will inherently preserve the provenance of the data records cryptographically, making it impossible to deny the misuse of the data. 
There are already proposals for programmable DLT platforms that enable smart, contract-based programming models for decentralized AI applications, which ensure the self-execution of AI agents based on predefined terms and conditions that will ultimately lead to innovations in systems suited for explainable AI~\cite{marwala2018blockchain}.

Undoubtedly, it is desirable to have an immutable trail to track the development of the data flow and complex behaviors of AI-based systems for model debugging purposes. 
DLTs can do precisely that, tracking every step in the data processing and decision-making chain. 
Through tracking behaviors of AI-based systems across different data input and application scenarios, we gain more understanding of and confidence in the decisions made by those systems.  
Furthermore, it provides insights into tuning those ``black boxes" to balance performance and prediction accuracy with the explainability of the system. 
In case of unfortunate and/or unforeseen incidents that arise due to the application of the AI models, these DLT-based trails will be essential to determine whether humans (and who precisely) or machines are at fault ~\cite{8481263}. If it is later discovered that a dataset is corrupted long after the model is trained, it may be hard to figure out which estimated parameters have been corrupted as well, and the influence of the corrupted data on the model output. 
However, according to Marechaux et al.~\cite{marechauxtowards}, if AI training is treated as a DLT transaction, then the ledger will store valuable traceability information. 
During model training, a transaction record is created to store contextual information in the ledger, such as the training model type, the dataset used, and the value of the parameters, both before and after the training. 
Therefore, DLTs can be leveraged to improve dataset traceability and consistency of ML models.

There are several notable applications of DLT in AI to provide explainability.
Salah et al. identify that DLTs can help in designing trustworthy and interpretable transparent AI algorithms to know why the algorithm is reaching a specific decision by tracing executions in many application areas, including healthcare, military, and autonomous vehicles~\cite{8598784}. 
Nassar et al. propose a model in which AI and explainable AI nodes or predictors that act as trusted oracles, perform computation, and interact with smart contracts deployed on DLT-systems which record and log execution outcomes and decisions in the immutable ledger~\cite{nassarblockchain}. 
Ferrer et al. have explored a future in which untrusting devices, for example, swarm robotics, Internet of Thing  (IoT) devices, or cell phones, will coordinate and make joint decisions~\cite{ferrer2018blockchain}, and the ledger will be used to explain the decisions of the collective AI agents, after-the-fact.

Recently deep fake images and videos have seen an uptick in contributing to misinformation on the Web. 
In order to address this problem Hassan et al. propose a DLT-based solution for proof of authenticity of digital videos in which a secure and trusted traceability to the original video creator or source can be established, in a decentralized manner. 
Their solution makes use of a decentralized storage system called the Inter-Planetary File System (IPFS), Ethereum name service, and a decentralized reputation system.  
Their premise is that if a video or piece of digital content is not traceable, then the digital content cannot be trusted~\cite{hasan2019combating}. 
The digital trace, or the lack of a digital trace, provides the explanation for the deep fake content.
Calvaresi et al. describe a system that combines explainable AI, with DLT to ensure trust in domains where, due to environmental constraints or to some characteristics of the users/agents in the system, the effectiveness of the explanation may drop dramatically~\cite{calvaresi2019explainable}. 
They draw an example from Unmanned Aerial Vehicles (UAVs)  working in a multi-agent autonomous system. 

\section{Conclusion} \label{chapter41-conclusions}
We have reviewed and summarized approaches from research that we believe will serve as directions for explainable AI. With the increasing focus on explainable AI, we are at the cusp of a new era of AI, where explainability plays a pivotal role in the adoption of AI systems. This renewed interest has resulted in review papers and position statements that call for greater user-centric explainability. From our literature review and our resulting set of synthesized definitions of explanations and explainable knowledge-enabled systems, we have identified current directions for research, in addition to building a catalog of hybrid explanations, that will contribute in different ways to provide ``user-comprehensible, context-aware, and provenance-enabled" explanations.

In our previous review (``Foundations of Explainable Knowledge-enabled Systems") of AI systems, we showed how different AI domains (i.e., expert systems, Semantic Web, cognitive assistants, and ML domains) and varying methodologies are suited to different aspects of explanations. In this chapter, we have built on this review and have identified AI methods that will aid in improving particular aspects of explainability, such as trust, comprehensibility, adaptiveness and causality. The next-generation hybrid AI systems would benefit from these identified strengths, utilizing a (potentially carefully chosen) collection of these techniques in combination to provide more complete and satisfying explanations. More specifically, we have shown that causal and neuro-symbolic methods, semantic representations of explainability, and mechanisms for trustworthy knowledge sharing through DLTs may play important roles in the future.

Further, we noted that different situations, contexts, and user requirements demand explanations of varying complexities, granularities, levels of evidence, presentations, etc. We presented a catalog of the nine explanation types that we synthesized from our literature review and our work on explaining complex and customized health assistants.
We believe these explanation descriptions may support an explanation methodology style and even a technical selection approach when designing customized explanation components. 

In conclusion, we believe that the increased adoption of AI systems and the need not only to understand the rationale behind their decisions, but also to make the results more useful in context and helpful in furthering joint human/computer reasoning, will lend to the development of new, explainable AI techniques. Furthermore, we opine that these explainable AI techniques will leverage different knowledge silos and sources, utilize a combination of explanation types, and incorporate mechanisms to improve the interpretability of AI models. We believe that current research and future research in explainable knowledge-enabled systems will serve as a means to build a more comprehensive and user-centric understanding of explainability.   

\section{Acknowledgments}
This work is partially supported by IBM Research AI through the AI Horizons Network. We thank our colleagues from IBM Research, Amar Das, Morgan Foreman and Ching-Hua Chen, and from RPI, James P. McCusker, and Rebecca Cowan, who greatly assisted the research and document preparation.
\makeatother
\bibliographystyle{IEEETrans}
\bibliography{4-charigruensenevomcguinness}

\end{document}